\documentclass{article}

\PassOptionsToPackage{numbers, compress}{natbib}
\usepackage[preprint]{neurips_2024}

\usepackage[utf8]{inputenc} %
\usepackage[T1]{fontenc}    %
\usepackage{url}            %
\usepackage{booktabs}       %
\usepackage{amsfonts}       %
\usepackage{nicefrac}       %
\usepackage{microtype}      %
\usepackage{xcolor}         %

\usepackage{graphicx}
\usepackage{xcolor} %
\usepackage{todonotes}
\usepackage{subcaption}

\usepackage{amsmath}
\usepackage{amssymb}
\usepackage{mathtools}
\usepackage{amsthm}

\usepackage{algcompatible}
\usepackage{algorithm}

\usepackage{listings}

\usepackage{hyperref}       %
\hypersetup{colorlinks=true,allcolors=black}

\usepackage[capitalize,nameinlink]{cleveref}

\theoremstyle{plain}
\newtheorem{theorem}{Theorem}[section]

\theoremstyle{definition}
\newtheorem{definition}[theorem]{Definition}

\theoremstyle{remark}

\renewcommand{\paragraph}[1]{\noindent\textbf{#1}~~}

\title{iQRL -- Implicitly Quantized Representations for Sample-efficient Reinforcement Learning}
\author{%
  Aidan Scannell \\
  Aalto University\\
  \texttt{\small aidan.scannell@aalto.fi}\\
  \And
  Kalle Kujanpää \\
  Aalto University\\
  \texttt{\small kalle.kujanpaa@aalto.fi}\\
  \And
  Yi Zhao \\
  Aalto University\\
  \texttt{\small yi.zhao@aalto.fi}\\
  \And
  Mohammadreza Nakhaei\\
  Aalto University\\
  \texttt{\small mohammadreza.nakhaei@aalto.fi}\\
  \And
  Arno Solin \\
  Aalto University\\
  \texttt{\small arno.solin@aalto.fi}\\
  \And
  Joni Pajarinen \\
  Aalto University\\
  \texttt{\small joni.pajarinen@aalto.fi}\\
}

\newcommand{\our}{\textsc{iQRL}\xspace}

\usepackage{xspace}
\newcommand{\eg}{\textit{e.g.\@}\xspace}
\newcommand{\ie}{\textit{i.e.\@}\xspace}

\usepackage{bm}

\def\va{{\bm{a}}}

\def\vo{{\bm{o}}}

\def\vq{{\bm{q}}}

\def\vv{{\bm{v}}}

\def\vx{{\bm{x}}}

\def\vz{{\bm{z}}}

\def\mA{{\bm{A}}}

\newcommand{\E}{\mathbb{E}}

\newcommand{\R}{\mathbb{R}}

\usetikzlibrary{shapes.geometric,positioning,calc}

  \newlength{\nodedist}
  \newlength{\nodedistv}

\definecolor{cube}{HTML}{f7f75e}
\newcommand{\cube}{%
\resizebox{1.5em}{1.4em}{
\begin{tikzpicture}[baseline=-1ex,cube/.style={very thick,black},
            grid/.style={very thin,gray},
            axis/.style={->,black,thick}]
 \begin{scope}[every node/.append style={yslant=-0.5},yslant=-0.5]
 [cube/.style={very thick,black},
            axis/.style={->,blue,thick}]
   \shade[right color=cube, left color=black!50!cube] (0,0) rectangle +(3,3);
   \node at (0.5,2.5) {};
   \node at (1.5,2.5) {};
   \node at (2.5,2.5) {};
   \node at (0.5,1.5) {};
   \node at (1.5,1.5) {};
   \node at (2.5,1.5) {};
   \node at (0.5,0.5) {};
   \node at (1.5,0.5) {};
   \node at (2.5,0.5) {};
   \draw (0,0) grid (3,3);
 \end{scope}

 \begin{scope}[every node/.append style={yslant=0.5},yslant=0.5]
   \shade[right color=gray!70,left color=cube] (3,-3) rectangle +(3,3);
   \node at (3.5,-0.5) {};
   \node at (4.5,-0.5) {};
   \node at (5.5,-0.5) {};
   \node at (3.5,-1.5) {};
   \node at (4.5,-1.5) {};
   \node at (5.5,-1.5) {};
   \node at (3.5,-2.5) {};
   \node at (4.5,-2.5) {};
   \node at (5.5,-2.5) {};
   \draw (3,-3) grid (6,0);
 \end{scope}

 \begin{scope}[every node/.append style={
     yslant=0.5,xslant=-1},yslant=0.5,xslant=-1
   ]
   \shade[bottom color=cube, top color=black!80] (6,3) rectangle +(-3,-3);
   \node at (3.5,2.5) {};
   \node at (3.5,1.5) {};
   \node at (3.5,0.5) {};
   \node at (4.5,2.5) {};
   \node at (4.5,1.5) {};
   \node at (4.5,0.5) {};
   \node at (5.5,2.5) {};
   \node at (5.5,1.5) {};
   \node at (5.5,0.5) {};
   \draw (3,0) grid (6,3);
 \end{scope}
\end{tikzpicture}}}

\begin{document}

\maketitle

\begin{abstract}
  Learning representations for reinforcement learning (RL) has shown much promise for continuous control. We propose an efficient representation learning method using only a self-supervised latent-state consistency loss. Our approach employs an encoder and a dynamics model to map observations to latent states and predict future latent states, respectively. We achieve high performance and prevent representation collapse by quantizing the latent representation such that the rank of the representation is empirically preserved. Our method, named iQRL: \textbf{i}mplicitly \textbf{Q}uantized \textbf{R}einforcement \textbf{L}earning, is straightforward, compatible with any model-free RL algorithm, and demonstrates excellent performance by outperforming other recently proposed representation learning methods in continuous control benchmarks from DeepMind Control Suite.
\end{abstract}

\section{Introduction}
\label{intro}

Reinforcement learning (RL, \eg, \cite{sutton2018reinforcement}) has shown much promise for solving complex continuous
control tasks.
However, applying RL in real-world environments is challenging as it typically requires millions of data points
which can be unpractical---\ie RL is sample inefficient.
On the other hand, representation learning has become a widely adopted solution for improving sample efficiency in deep learning.
The core idea is to learn features which capture the underlying structure and patterns of the data.
In the context of RL, such features can be learned independently from the downstream task.
Whilst representation learning has had successes in RL, these have mainly been restricted to image-based observations (\eg,
CURL~\cite{laskinCURLContrastiveUnsupervised2020}, DrQ~\cite{yaratsImageAugmentationAll2020},
DrQ-v2~\cite{yaratsMasteringVisualContinuous2021}, and TACO~\cite{zhengTextttTACOTemporal2023}).

The investigation of representation learning for state-based RL is much less common.
This is likely due to the fact that learning a compact representation of an already compact state vector seems unnecessary.
However, recent work by \citet{fujimotoSALEStateActionRepresentation2023,zhaoSimplifiedTemporalConsistency2023}
suggests that the difficulty of a task is due to the complexity of the underlying transition dynamics,
as opposed to the size of the observation space.
As such, investigating representation learning for state-based RL is a promising research direction.

Recently, TCRL \cite{zhaoSimplifiedTemporalConsistency2023}
and SPR \cite{schwarzerDataEfficientReinforcementLearning2020}
have obtained state-of-the-art performance on continuous control benchmarks by learning
representations with self-supervised losses.
Self-supervised learning (SSL) approaches (which do not reconstruct observations) attempt to learn good features
without labels \cite{anandUnsupervisedStateRepresentation2019}.
Whilst they can learn robust representations, self-supervised losses are susceptible to a problem known
as representation collapse
(see \cref{def:complete-collapse}),
where the encoder learns to map all observations to a constant latent
representation \citep{jingUnderstandingDimensionalCollapse2021}.
As such, when leveraging SSL approaches to learn representations for RL,
it is common to combine the self-supervised latent-state consistency loss with
other loss terms, such as minimizing the reward prediction error in the latent space
\citep{zhangLearningInvariantRepresentations2020,zhaoSimplifiedTemporalConsistency2023,hansenTemporalDifferenceLearning2022,geladaDeepMDPLearningContinuous2019,rezaei-shoshtariContinuousMDPHomomorphisms2022}.
This helps to prevent representation collapse at the cost of learning a task-specific representation.

\begin{figure}[t!]
  \small

  \definecolor{color0}{HTML}{f8f7ea}
  \definecolor{color1}{HTML}{7cf675}
  \definecolor{color2}{HTML}{f7f75e}
  \definecolor{color3}{HTML}{e27b75}
  \definecolor{color4}{HTML}{f7c075}
  \definecolor{color5}{HTML}{7cf6ea}
  \definecolor{color6}{HTML}{e2bdea}

  \newcommand{\codebook}[1]{\tikz[baseline=-.5ex]{%
    \foreach \c [count=\i] in {1,...,#1}
      \node[fill=color\c,minimum height=5pt,minimum width=5pt,inner sep=0,draw=black!80] at (\i*5pt,0) {};
  }}  
  
  \resizebox{\textwidth}{!}{%
  \begin{tikzpicture}

  \pgfdeclarelayer{background}
  \pgfsetlayers{background,main}

  \newcommand{\sub}[0]{\scalebox{.8}{\ensuremath t}}
  \newcommand{\subs}[1]{\scalebox{.8}{\ensuremath t{+}#1}}

  \tikzstyle{mynode}=[fill=black!5,draw=black!80,rounded corners=1pt,font=\scriptsize,inner sep=0,align=center,line width=.6pt]
  \tikzstyle{trap}=[mynode,fill=color1,trapezium,text width=4em, minimum height=1em,
                    trapezium left angle=-70, trapezium right angle=-70,inner sep=2pt]
  \tikzstyle{blob}=[mynode,circle,minimum width=2.2em,minimum height=2.2em]
  \tikzstyle{polval}=[mynode,minimum width=6em,minimum height=2em,align=center,text=white,fill=black!60,inner sep=2pt]
  \tikzstyle{arr}=[line width=1pt,black,->]
  \tikzstyle{darr}=[line width=1pt,black,<->,densely dotted]
  \tikzstyle{dlc}=[align=center,font=\scriptsize,text width=5em]

  \draw[fill=color0,draw=color0!60!black,rounded corners=4pt] (-.5\nodedist,-.7\nodedistv) rectangle (2.5\nodedist,1.5\nodedistv);

  \node[blob,fill=color3] (z0) at (0\nodedist,0) {$\vz_{\sub}\vphantom{\hat\vz_{\subs{1}}}$};
  \node[blob,fill=color4] (z1) at (1\nodedist,0) {$\hat\vz_{\subs{1}}$};
  \node[blob,fill=color4] (z2) at (2\nodedist,0) {$\hat\vz_{\subs{2}}$};

  \node[trap] (e0) at (0\nodedist,\nodedistv) {Encoder};%
  \node[trap] (e1) at (1\nodedist,\nodedistv) {EMA enc.};%
  \node[trap] (e2) at (2\nodedist,\nodedistv) {EMA enc.};%

  \node[blob,fill=color3] (zh1) at ($(z1)!0.5!(e1)$) {$\bar\vz_{\subs{1}}$};
  \node[blob,fill=color3] (zh2) at ($(z2)!0.5!(e2)$) {$\bar\vz_{\subs{2}}$};

  \coordinate (m0) at ($(z0)!.5!(z1)$);
  \node[blob,fill=color6] (a0) at ($(m0)!(zh1)!(m0)$) {$\va_{\sub}$};
  \coordinate (m1) at ($(z1)!.5!(z2)$);
  \node[blob,fill=color6] (a1) at ($(m1)!(zh2)!(m1)$) {$\va_{\subs{1}}$};

  \node[mynode,minimum width=5em,minimum height=5em,outer sep=0,node distance=6em,path picture={\node at (path picture bounding box.center){\includegraphics[height=7em]{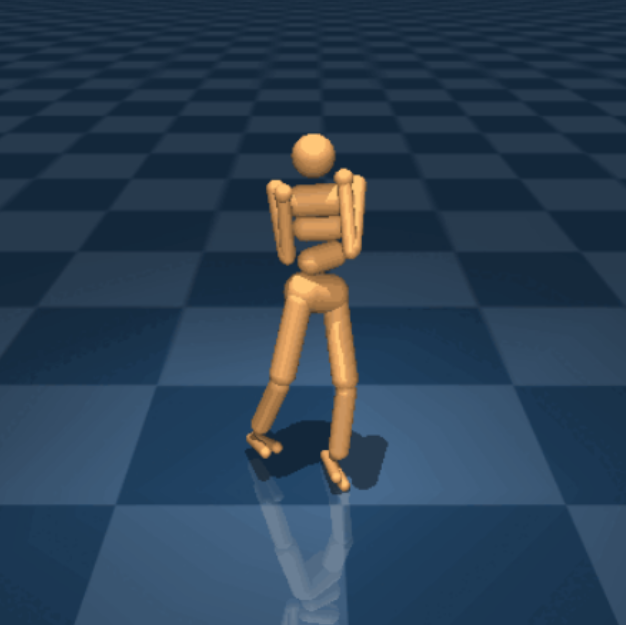}};}] (i0) at (0\nodedist,1.4\nodedistv) {};

  \node[mynode,minimum width=5em,minimum height=5em,outer sep=0,node distance=6em,path picture={\node at (path picture bounding box.center){\includegraphics[height=7em]{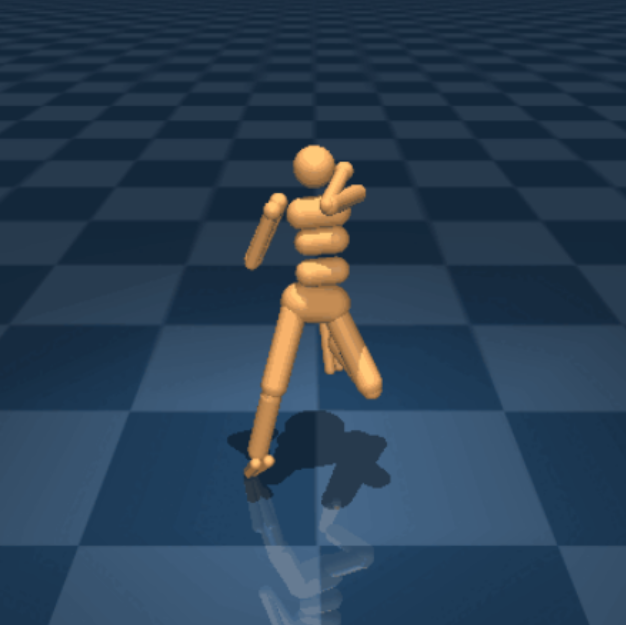}};}] (i1) at (1\nodedist,1.4\nodedistv) {};

  \node[mynode,minimum width=5em,minimum height=5em,outer sep=0,node distance=6em,path picture={\node at (path picture bounding box.center){\includegraphics[height=7em]{figs/humanoid-3}};}] (i2) at (2\nodedist,1.4\nodedistv) {};

  \node[anchor=north] (o0) at (i0.north) {\color{white}$o_{\sub}$};
  \node[anchor=north] (o1) at (i1.north) {\color{white}$o_{\subs{1}}$};
  \node[anchor=north] (o2) at (i2.north) {\color{white}$o_{\subs{2}}$};

  \draw[arr] (i0) -- (e0);
  \draw[arr] (i1) -- (e1);
  \draw[arr] (i2) -- (e2);
  \draw[arr] (z0) -- node[below,font=\scriptsize,xshift=-1em] {Dynamics} (z1);
  \draw[arr] (z1) -- node[below,font=\scriptsize,xshift=-1em] {Dynamics} (z2);
  \draw[arr] (e0) -- node[above,font=\scriptsize,rotate=90,yshift=6pt] {FSQ} (z0);
  \draw[arr] (e1) -- (zh1);
  \draw[arr] (e2) -- (zh2);
  \draw[arr] (a0) -- (z1);
  \draw[arr] (a1) -- (z2);
  \draw[darr] (zh1) -- (z1);
  \draw[darr] (zh2) -- (z2);
  \draw[arr,-] (z2) -- ($(e2.east)!(z2)!(e2.east)$);

  \node[circle,fill=color0,inner sep=0, scale=.7] at ($(e0)!.5!(z0)$) {\cube};
  \node[circle,fill=color0,inner sep=0, scale=.7] at ($(e1)!.4!(zh1)$) {\cube};
  \node[circle,fill=color0,inner sep=0, scale=.7] at ($(e2)!.4!(zh2)$) {\cube};
  \node[circle,fill=color0,inner sep=0, scale=.7] at ($(z0)!.7!(z1)$) {\cube};
  \node[circle,fill=color0,inner sep=0, scale=.7] at ($(z1)!.7!(z2)$) {\cube};

  \node[dlc] (cb0) at (0\nodedist,-.5\nodedistv) {\codebook{5} \\ Discrete latent codes};
  \node[dlc] (cb1) at (1\nodedist,-.5\nodedistv) {\codebook{5} \\ Discrete latent codes};
  \node[dlc] (cb2) at (2\nodedist,-.5\nodedistv) {\codebook{5} \\ Discrete latent codes};

  \draw[->, thick, bend left=45,shorten >=0pt, shorten <=2pt] (z0) to (cb0);
  \draw[->, thick, bend left=45,shorten >=2pt, shorten <=0pt] (cb0) to (z0);

  \draw[->, thick, bend left=45,shorten >=0pt, shorten <=2pt] (z1) to (cb1);
  \draw[->, thick, bend left=45,shorten >=2pt, shorten <=0pt] (cb1) to (z1);

  \draw[->, thick, bend left=45,shorten >=0pt, shorten <=2pt] (z2) to (cb2);
  \draw[->, thick, bend left=45,shorten >=2pt, shorten <=0pt] (cb2) to (z2);

  \draw[fill=color0,draw=color0!60!black,rounded corners=4pt] (2.7\nodedist,-.7\nodedistv) rectangle (5.0\nodedist,1.5\nodedistv);

  \node[mynode,minimum width=5em,minimum height=5em,outer sep=0,node distance=6em,path picture={\node at (path picture bounding box.center){\includegraphics[height=7em]{figs/humanoid-1}};}] (i3) at (3.2\nodedist,1.4\nodedistv) {};

  \node[mynode,minimum width=5em,minimum height=5em,outer sep=0,node distance=6em,path picture={\node at (path picture bounding box.center){\includegraphics[height=7em]{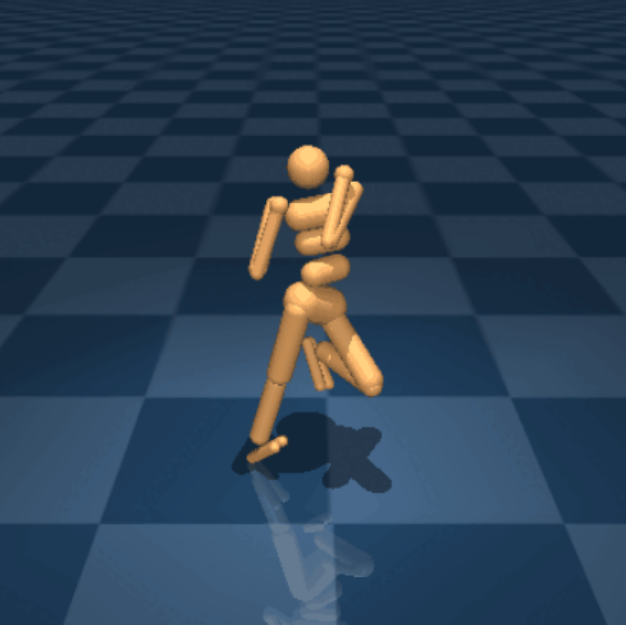}};}] (i4) at (4.2\nodedist,1.4\nodedistv) {};

  \node[anchor=north] (o3) at (i3.north) {\color{white}$o_{\sub}$};
  \node[anchor=north] (o4) at (i4.north) {\color{white}$o_{\subs{1}}$};

  \node[trap] (e3) at (3.2\nodedist,\nodedistv) {Encoder};
  \node[trap] (e4) at (4.2\nodedist,\nodedistv) {Encoder};

  \node[blob,fill=color3] (z3) at (3.2\nodedist,.25\nodedistv) {$\vz_{\sub}\vphantom{\hat\vz_{\subs{1}}}$};
  \node[blob,fill=color3] (z4) at (4.2\nodedist,.25\nodedistv) {$\vz_{\subs{1}}$};
  \node[blob,fill=color6] (a5) at (3.7\nodedist,.25\nodedistv) {$\va_{\sub}$};
  \node[blob,fill=color2] (a4) at (4.7\nodedist,.25\nodedistv) {$r_{\subs{1}}$};

  \draw[arr] (i3) -- (e3);
  \draw[arr] (i4) -- (e4);
  \draw[arr] (e3) -- (z3);
  \draw[arr] (e4) -- (z4);

  \node[circle,fill=color0,inner sep=0, scale=.7] at ($(e3)!.4!(z3)$) {\cube};
  \node[circle,fill=color0,inner sep=0, scale=.7] at ($(e4)!.4!(z4)$) {\cube};

  \draw[draw=black!80,rounded corners=2pt] ($(z3) - (2em,2.5em)$) rectangle ($(a4) + (2em,2em)$);
  \node[anchor=south west,font=\scriptsize] at ($(z3) + (-2em,-2.5em)$) {Latent transition};

  \node[draw=black!80,rounded corners=2pt,align=center] (rl) at (3.95\nodedist,-.4\nodedistv) {
  \begin{minipage}{3cm}
    \vspace*{-6pt}
    \begin{align*}
      \text{Actor:}\quad & \pi(\vz_{\sub}) \\
      \text{Critic:}\quad & Q(\vz_{\sub},\va_{\sub})
    \end{align*}
  \end{minipage}};
  
  \node[anchor=south,font=\scriptsize] at (rl.north) {Any model-free RL algorithm};
  
\end{tikzpicture}}
  \caption{\textbf{Overview.} \our is a stand-alone representation learning technique that is compatible with any model-free RL algorithm (we use TD3 \citep{fujimotoAddressingFunctionApproximation2018}). Importantly, \our quantizes the latent representation with Finite Scalar Quantization (FSQ, \scalebox{.5}{\protect\cube}), using only a self-supervised latent-state consistency loss, \ie no decoder (see \cref{eq:rep-loss}). Making the latent representation discrete with an implicit codebook (\protect\codebook{6}) contributes to the very high sample efficiency of \our and empirically prevents representation collapse. Thanks to the FSQ-based quantization, \our does not need a reward prediction head to prevent representation collapse, a well-known issue with self-supervised learning, making the representation task-agnostic.} %
  \label{fig:overview}
\end{figure}

In this paper, we propose a simple representation learning technique which learns a task-agnostic representation
using only a self-supervised loss.
It is based solely on the latent-state consistency loss, \ie a commonly used self-supervised loss for continuous RL.
Importantly, our method empirically prevents representation collapse as it preserves the rank of the representation.
We accomplish this by quantizing our latent representation
with Finite Scalar Quantization \citep{mentzerFiniteScalarQuantization2023}, without using any reconstruction loss.
As a result, our latent space is bounded and associated with an \emph{implicit} codebook, whose size we can control.
Our method can be combined with any model-free RL method (we use TD3, \cite{fujimotoAddressingFunctionApproximation2018}).
See \cref{fig:overview} for an overview of our representation learning method.
Importantly, our method {\em (i)}~alleviates representation collapse,
{\em (ii)}~demonstrates excellent sample efficiency outperforming TCRL and TD7 on a wide range of different continuous control tasks,
{\em (iii)}~is simple to implement, and {\em (iv)}~learns a task-agnostic
representation that could be helpful in downstream tasks.

\section{Related Work}
\label{sec:related_work}
In this section, we recap methods for representation learning in RL.
In particular, we motivate why researchers are moving towards learning representations using self-supervised learning.
Then, as our method builds upon self-supervised representation learning, which is susceptible to representation
and dimensional collapse (see \cref{def:complete-collapse,def:dim-collapse}),
we review contrastive self-supervised representation learning approaches; an alternative approach to
preventing representation collapse.

\paragraph{Representation learning}
Learning representations for RL has been investigated for decades
\citep{abelOptimalBehaviorApproximate2016,mannorDynamicAbstractionReinforcement2004,liUnifiedTheoryState2006,andreStateAbstractionProgrammable2002,deardenAbstractionApproximateDecisiontheoretic1997,singhReinforcementLearningSoft1994,higginsDefinitionDisentangledRepresentations2018,vanhoofStableReinforcementLearning2016,watterEmbedControlLocally2015,ghoshRepresentationsStableOffPolicy2020}.
However, these approaches are usually limited to simple environments.
More recently, \citet{fujimotoSALEStateActionRepresentation2023} proposed TD7, an extension of TD3 which learns
state and action embeddings and then performs TD3 with this representation, making it highly similar to our method, which also uses TD3 as the base algorithm.
However, their method uses a self-supervised loss with no explicit mechanism to prevent representation collapse.
In contrast to TD7 and motivated by representation collapse, we quantize our latent space, which we show empirically
prevents representation collapse.

\paragraph{Observation reconstruction}
A prominent idea in both model-based and model-free RL has been to learn latent representations
with reconstruction objectives (\eg\ VAE, \cite{kingmaAutoEncoding2014})
\citep{finnDeepSpatialAutoencoders2016,higginsDARLAImprovingZeroShot2017,langeAutonomousReinforcementLearning2012,watterEmbedControlLocally2015,hafner2019dream,rubinstein1997optimization}.
However, as these approaches use observation reconstruction to learn a representation
their latent representation contains information about the observation
which cannot be controlled by the agent and is not relevant for solving the
task, which distorts the optimization landscape \cite{zhang2018natural,zintgraf2021varibad}.
In our experiments, we show that learning representations with observation reconstruction for model-free RL
not only harms sample efficiency, but in the complex DMC Dog Run task, it can prevent the agent from solving the task.

\paragraph{Latent-state consistency}
How can effective representations be learned efficiently without resorting to reconstruction? A common solution has been to
attach an auxiliary loss to the RL objective and perform representation learning
\cite{tomar2021learning,niBridgingStateHistory2023}.
A promising approach for learning suitable representations is the use of self-predictive abstractions, where the model is trained to predict future latent states through an auxiliary loss \cite{subramanian2022approximate}.
\citet{ye2021mastering} introduce a self-supervised consistency loss on the learned latent representation.
Instead of relying on a reconstruction-based loss function, \citet{schwarzer2020data} propose a cosine similarity loss between predicted future latent states and the true future latent states and then perform Q-learning in the learned latent space.
Our approach shares similarities with SPR \citep{schwarzer2020data}, however, we focus on state-based observations,
which leads us to quantizing our representation to prevent representation collapse,
instead of using a projection head \cite{wenMechanismPredictionHead2022}.

\paragraph{Latent-state consistency for model-based RL}
Similarly to model-free RL, using the reconstruction loss for learning representation is also unreliable in model-based RL \cite{lutter2021learning} and can have a detrimental effect on the performance of model-based methods in various benchmarks \cite{kostrikov2020image, yarats2021mastering}.
Therefore, in the context of model-based RL,
TD-MPC/TD-MPC2 \cite{hansenTemporalDifferenceLearning2022,hansenTDMPC2ScalableRobust2023} use a consistency loss to learn representations for planning with Model Predictive Path Integral control together with reward and value functions learned through temporal difference methods \cite{williams2015model}.
\citet{zhaoSimplifiedTemporalConsistency2023} show that the planning component of TD-MPC is not strictly necessary for high
performance and applying model-free RL on top of the self-consistent representations is sufficient for
performance competitive with state-of-the-art.
We build on top of TCRL to show that we can combat representation collapse through latent-space quantization.
As a result, we can drop the reward prediction head to learn a task-agnostic representation.\looseness-1

\paragraph{Contrastive learning}
An alternative approach to preventing representation collapse in self-supervised learning is to use contrastive losses.
In the context of RL, this was done by CURL \citep{laskinCURLContrastiveUnsupervised2020}
and TACO \citep{zhengTextttTACOTemporal2023}.
Whilst CURL and TACO are designed for image-based observations, their contrastive learning approaches could still hold value
in state-based RL.
The main idea in contrastive learning is to prevent representation collapse (see \cref{def:complete-collapse})
by pushing the latent vectors associated with different observations away from each other.
Nevertheless, contrastive methods still experience dimensional collapse \citep{jingUnderstandingDimensionalCollapse2021}.
In contrast to TACO, we do not use a contrastive loss and instead leverage quantization to help prevent both representation
and dimensional collapse.
To offer a fair comparison between contrastive learning and our quantization scheme, we compare our method against a version of TACO
tuned for state-based RL.

\section{Preliminaries}
In this section, we introduce and formally define representation collapse.

\paragraph{Representation collapse}
Self-supervised learning methods learn representations by minimizing distances between two embedding vectors.
As such, there is a trivial solution where the encoder outputs a constant for all inputs.
Formally, this can be defined as follows:
\begin{definition}[Complete representation collapse] \label{def:complete-collapse}
  Given an encoder $e_{\theta}: \mathcal{O} \rightarrow \mathcal{Z}$ which maps observations $\vo \in \mathcal{O}$ to latent states $\vz \in \mathcal{Z}$, the representation is said to be completely collapsed when the latent representation is constant for all observations, \ie, $e_{\theta}(\vo) = c, \forall \vo \in \mathcal{O}$.
\end{definition}
In the context of SSL, \citet{jingUnderstandingDimensionalCollapse2021} investigated
another type of representation collapse known as dimensional collapse, which is defined as follows:
\begin{definition}[Dimensional collapse] \label{def:dim-collapse}
  Given an encoder $e_{\theta}: \mathcal{O} \rightarrow \mathcal{Z}$ which maps observations $\vo \in \mathcal{O}$ to latent states $\vz_{t} \in \mathcal{Z} =\R^{d}$, of dimension $d$, the representation is said to be dimensionally collapsed when the latent vectors only span a lower dimensional space.%
\end{definition}

\begin{figure*}[t]
\vskip 0.2in
\begin{center}
\centerline{\includegraphics[width=1.0\textwidth]{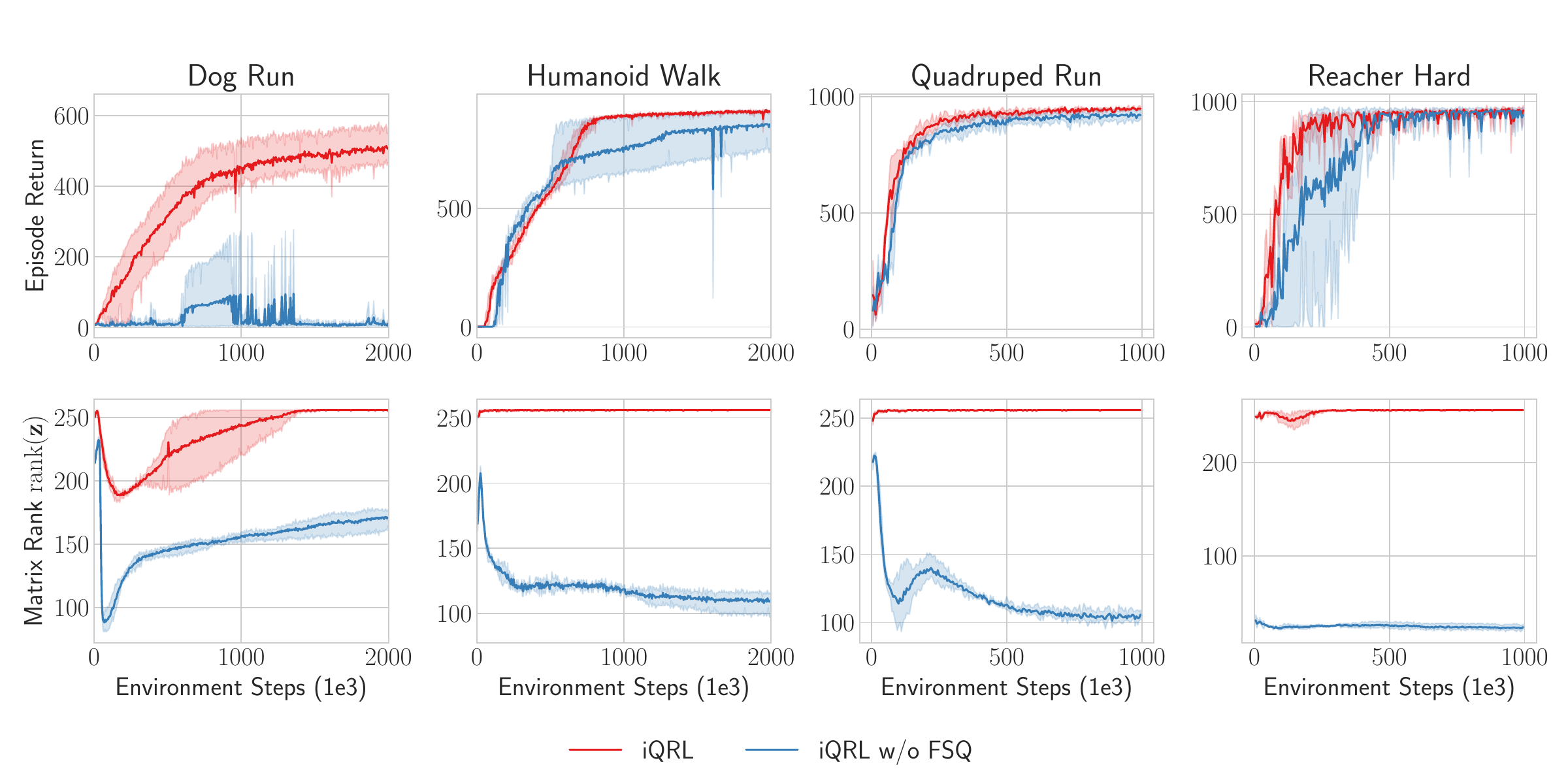}}
\caption{\textbf{Ablation of quantization.} We show how our quantization scheme prevents dimensional collapse. In all tasks, our FSQ scheme prevents dimensional collapse (red)  as the rank of the representation remains high. In contrast, when our quantization is not used (blue) the representation undergoes dimensional collapse, indicated by the rank reducing. In the Dog Run task, this results in the agent not learning to solve the task. We plot the mean (solid line) and the $95\%$ confidence intervals (shaded) across 5 random seeds, where each seed averages over 10 evaluation episodes.}
\label{fig:normalization-ablation}
\end{center}
\end{figure*}

Whilst complete representation collapse is a clear issue when learning representations for RL, it is not immediately obvious if dimensional collapse is an issue because the goal of representation learning is often considered to be learning a lower-dimensional representation of the observations. In other words, there is a trade-off: We want to learn a lower-dimensional representation, but at the same time, we want to ensure that the representation contains all the information required for predicting future states and, thus, state values \cite{niBridgingStateHistory2023}. Our experiments show that whilst dimensional collapse is not always an issue, in some more complex environments, it can prevent agents from learning to solve a task (see \cref{fig:normalization-ablation}).

\begin{algorithm}[tb]
   \caption{\our}
   \label{alg:main_alg}
   \renewcommand{\algorithmiccomment}[1]{\hfill\textcolor{gray}{\(\triangleright\) #1}}
\begin{algorithmic}
   \STATE {\bfseries Input:} Encoder $e_{\theta}$, dynamics $d_{\phi}$, critics $\{q_{\psi_{1}}, q_{\psi_{2}} \}$, policy $\pi_{\eta}$, learning rate $\alpha$, target network update rate $\tau$
   \FOR{$i$ {\bfseries to} $N_{\text{episodes}}$}
    \STATE $\mathcal{D} \leftarrow \mathcal{D} \cup \{\vo_{t}, \va_{t}, \vo_{t+1}, r_{t+1}\}^{T}_{t=0}$ \COMMENT{Collect data in environment}
    \FOR{$i=1$ {\bfseries to} $T$}
        \STATE $[\theta, \phi] \leftarrow [\theta, \phi] + \alpha \nabla \left( \mathcal{L}_{\text{rep}}(\theta, \phi; \mathcal{D}) \right)$  \COMMENT{pdate representation, \cref{eq:rep-loss}}
        \STATE $\psi \leftarrow \psi + \alpha \nabla \left( \mathcal{L}_{q}(\psi; \mathcal{D}) \right)$ \COMMENT{Update critic, \cref{eq:value-loss}}
        \IF{$i$ \% 2 == 0}
          \STATE $\eta \leftarrow \eta + \alpha \nabla \left( \mathcal{L}_{\pi}(\eta; \mathcal{D}) \right)$  \COMMENT{Update actor less frequently than critic, \cref{eq:policy-loss}}
        \ENDIF
        \STATE $ [\bar{\theta}, \bar{\psi}, \bar{\eta}] \leftarrow (1-\tau)  [\bar{\theta}, \bar{\psi}, \bar{\eta}] + \tau [{\theta}, {\psi}, \eta]$ \COMMENT{Update target networks}
    \ENDFOR
   \ENDFOR
\end{algorithmic}
\end{algorithm}

\section{Method}
\label{sec:method}

In this section, we detail our method, named \textit{implicitly Quantized Reinforcement Learning} (\our).
\our is conceptually simple, it {\em (i)}~learns a representation of the observation space and then,
{\em (ii)}~performs model-free RL (\eg, TD3) on this representation.
See \cref{fig:overview} and \cref{alg:main_alg}.

We consider Markov Decision Processes (MDPs, \cite{bellmanMarkovianDecisionProcess1957a}) $\mathcal{M} = (\mathcal{O}, \mathcal{A}, \mathcal{P}, \mathcal{R}, \gamma)$,
where an agent receives an observation $\vo_{t} \in \mathcal{O}$ at time step $t$, performs an action $\va_{t} \in \mathcal{A}$, and
then obtains the next observation $\vo_{t+1} = \mathcal{P}(\cdot \mid \vo_{t}, \va_{t})$ and reward $r_{t} = \mathcal{R} (\vo_{t}, \va_{t})$.
The discount factor is denoted $\gamma \in [0, 1)$.\looseness-4

\paragraph{Method components}
\our has four main components which we wish to learn:
\begin{align}
&\text{Encoder: } & \vz_{t} &= f(e_{\theta} (\vo_{t})) \label{eq:encoder} \\
&\text{Dynamics: } & \hat{\vz}_{t+1} &= {f}({\vz}_{t} + d_{\phi} (\vz_{t}, \va_{t})) \label{eq:transition} \\
&\text{Value: } & \vq_{t} &= \mathbf{q}_{\psi} (\vz_{t}, \va_{t}) \label{eq:value} \\
&\text{Policy: } & \va_{t} &\sim \pi_{\eta} (\vz_{t}) \label{eq:policy}
\end{align}
The encoder $e_{\theta}$ and latent-space dynamics model $d_{\phi}$ are responsible for representation learning.
$f(\cdot)$ denotes our quantization scheme, which implicitly quantizes our latent representation (more details to follow).
The encoder (with quantization) $f \circ e_{\theta}(\cdot)$ maps observations $\vo_{t}$ to latent states $\vz_{t}$ and is responsible for learning a representation
which can aid RL.
The latent-space dynamics model (with quantization) $d_{\phi}(\cdot)$ predicts the next latent states $\hat\vz_{t+1}$ given a latent state $\vz_{t}$ and an action $\va_{t}$.
Its sole purpose is to aid representation learning by making the latent states temporally consistent.
Note that we do not use it for model-based RL.
Once we have the representation learned by our encoder, we map all observations to the latent space and perform model-free RL
in this latent space.
Throughout this paper, we use Twin Delayed Deep Deterministic Policy Gradient
(TD3, \citep{fujimotoAddressingFunctionApproximation2018}) as the base algorithm.
It consists of two state-action value functions $\{q_{\psi_{1}},q_{\psi_{2}} \}$, known as critics, and a deterministic
actor $\pi_{\eta}$.
Following prior works \citep{yaratsMasteringVisualContinuous2021}, we use a linear exploration noise schedule
which decays from $1$ to $0.1$ during training.

\paragraph{Representation learning}
Our representation learning uses the latent-state consistency loss,
\begin{align} \label{eq:rep-loss}
  \mathcal{L}_{\text{rep}}(\theta, \phi; \tau)
&= \sum_{h=0}^{H-1} \gamma_{\text{rep}}^{h} \left(\frac{{f}(\hat\vz_{t+h} + d_{\phi}(\hat{\vz}_{t+h}, \va_{t+h}))}{\|{f}(\hat\vz_{t+h} + d_{\phi}(\hat{\vz}_{t+h}, \va_{t+h}))\|_{2}}\right)^{\top}
\left(\frac{{f}(e_{\bar{\theta}}(\vo_{t+h+1}))}{\|{f}(e_{\bar{\theta}}(\vo_{t+h+1}))\|_{2}} \right),
\end{align}
which minimizes the cosine similarity between the next state predicted by the dynamics model $\hat{\vz}_{t+1} = f(\hat\vz_{t} + d_{\phi}(\hat{\vz}_{t}, \va_{t}))$
and the next state predicted by the momentum encoder $\bar{\vz}_{t+1} = f(e_{\bar{\theta}}(\vo_{t+1}))$.
The latent states are obtained with multi-step predictions in the latent
space $\hat{\vz}_{t+1} = f(\hat\vz_{t}+ d_{\phi}(\hat{\vz}_{t}, \va_{t}))$.
The initial mapping to the latent space $\hat{\vz}_{0} = f(e_{\theta}(\vo_{0}))$ uses the online encoder which
is being trained jointly with the dynamics model $d_{\phi}(\hat{\vz_{t}}, \va_{t})$.
The target $e_{\bar{\theta}}(\vo_{t+1})$ is calculated with the momentum encoder which uses an exponential moving average (EMA)
of the encoder's weights $\bar{\theta} \leftarrow (1-\tau)\bar{\theta} + \tau \theta$.
The target network update rate is denoted $\tau$.
Note that we do not use reward or value prediction for learning our representation and as a result, our representation
is task-agnostic.

\paragraph{Quantization}
Motivated by preventing dimensional collapse we quantize our latent space following the approach from
Finite Scalar Quantization (FSQ, \cite{mentzerFiniteScalarQuantization2023}).
Their important observation is that carefully bounding each dimension gives rise to an {\em implicit}
codebook $\mathcal{C}$ of a chosen size $|\mathcal{C}|$.
Having requested a $d\text{-dimensional}$ latent space, \our configures the encoder to output $c$ channels per dimension
such that the representation from the encoder $\vx = e_{\theta}(\vo) \in \R^{d\times c}$
and the dynamics model $\hat{\vx} = {\vz} + d_{\phi} (\vz, \va) \in \R^{d \times c}$ are in $\R^{d \times c}$.
To quantize $\vx$ (and $\hat{\vx}$) into a finite set of codewords, we first apply a bounding function $f(\cdot)$ and then we round to integers.
Let us consider a single dimension $j$ of the encoder's output $\vv=[\vx]_{j,:} \in \R^{c}$ which consists of $c\text{-channels}$, and demonstrate
how it is quantized.
We follow FSQ and choose $f(\cdot)$ such that each entry in $\tilde{\vv} = \mathrm{round}(f(\vv))$
takes one of $L_{i}$ unique values,
\begin{align}
f : \vv \rightarrow \lfloor L_i/2 \rfloor \text{tanh}(\vv),
\end{align}
where $L_{i}$ is a hyperparameter for channel $i$, specified as FSQ levels $\mathcal{L}=\{L_{1},\ldots,L_{c}\}$.
This gives an entry in our codebook $\tilde\vv \in \mathcal{C}$, where the {\em implied} codebook
is given by the product of these per-channel codebook sets.
The vectors in $\mathcal{C}$ can be enumerated giving a bijection from any $\tilde{\vv}$ to an integer in $\{1,2,\ldots, L^{c}\}$.
As an example, in some of our experiments, we used $d=512$ latent dimensions each with $c=2$
channels consisting of 8 levels, \ie we used FSQ levels $\mathcal{L} = \{L_1=8, L_2=8\}$. This corresponds to a codebook of size $|\mathcal{C}| = \prod_{i=1}^c L_{i} = 8 \times 8 = 64 = 2^6$ for each dimension.

Note that this quantization requires a round operation. As such, to propagate gradients through the round operation
we use straight-through gradient estimation (STE).
This is easily accomplished in deep learning libraries using stop gradient $\mathrm{sg}$ as
$\mathrm{round\_ste}(x) : x \rightarrow x + \mathrm{sg}(\mathrm{round}(x)-x)$.
FSQ has the following hyperparameters: we must specify the number of channels $c$ and the number of levels per channel
$\mathcal{L} = \{L_{1},\ldots,L_{c}\}$.
\cref{tab:fsq-levels} shows the recommended number of channels and number of levels per channel to obtain codebooks of different sizes \citep{mentzerFiniteScalarQuantization2023}.
\begin{table}[h]
\caption{FSQ levels $\mathcal{L}$ to approximate different codebook sizes $|\mathcal{C}|$.}
\label{tab:fsq-levels}
\begin{center}
\begin{small}
\begin{sc}
\begin{tabular}{lccccc}
\toprule
Target size $|\mathcal{C}|$ & $2^{4}$ & $2^{6}$ & $2^{8}$ & $2^{9}$ & $2^{10}$ \\
\midrule
Proposed $\mathcal{L}$ & $\{5,3\}$ & $\{8,8\}$ & $\{8,6,5\}$ & $\{8,8,8\}$ & $\{8,5,5,5\}$ \\
\bottomrule
\end{tabular}
\end{sc}
\end{small}
\end{center}
\end{table}

In practice, we found codebooks of size $|\mathcal{C}|=2^{6}$ sufficient for all environments in the DeepMind Control suite.
However, for more complex environments we hypothesize that larger codebooks will be required.

\begin{figure*}[t]
\vskip 0.2in
\begin{center}
\centerline{\includegraphics[width=1.0\textwidth]{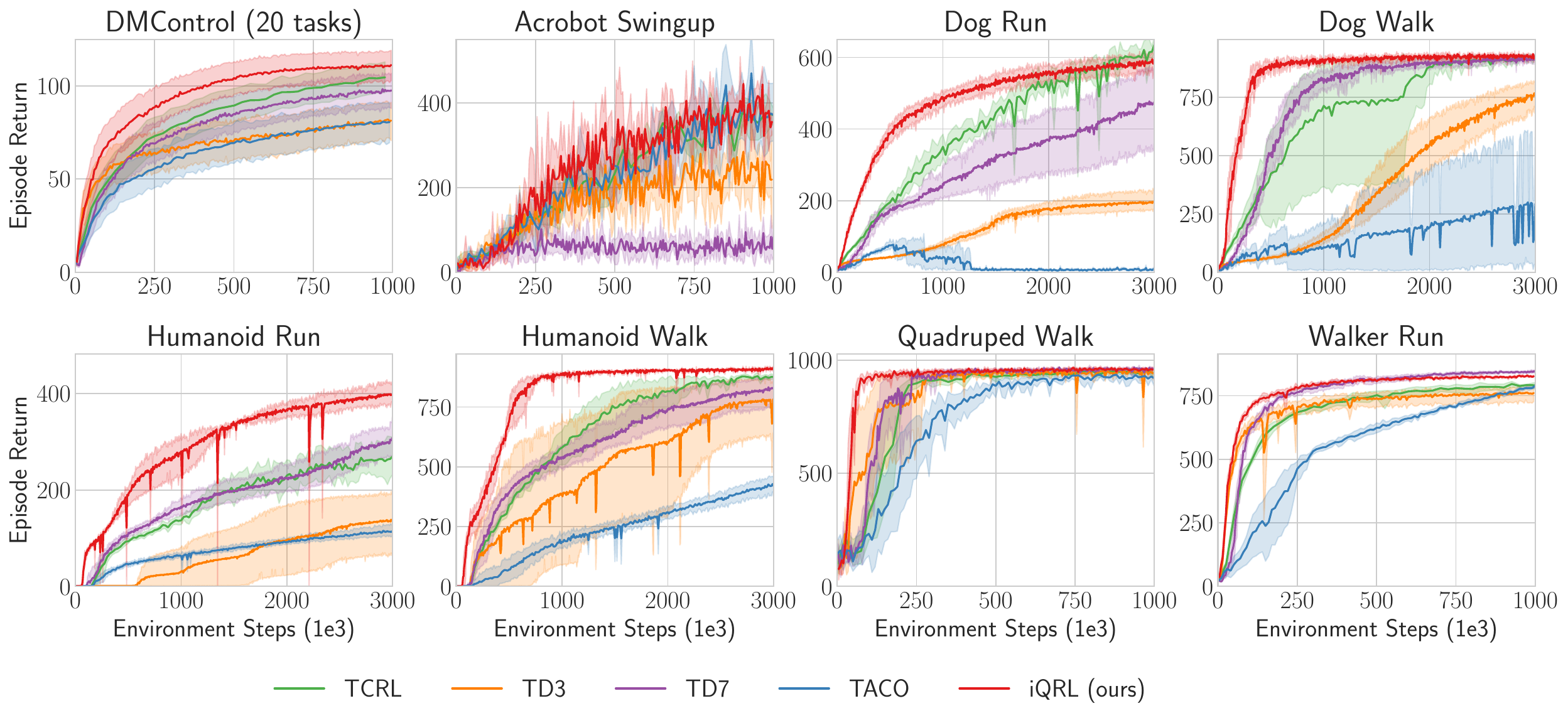}}

\caption{\textbf{DeepMind Control Suite results.} \our (red) is significantly more sample efficient than other model-free baselines TCRL (green), TD7 (purple), TACO (blue) and TD3 (orange). \our performs particularly well in the high-dimensional locomotion tasks and outperforms TCRL, which is the most similar baseline. Results are for 20 DMC tasks with UTD=1. We plot the mean (solid line) and the $95\%$ confidence intervals (shaded) across 5 random seeds, where each seed averages over 10 evaluation episodes. See \cref{fig:dmc_grid} for results in other DMC tasks.}
\label{fig:normalization_improves_sample_efficiency}
\end{center}
\end{figure*}

\paragraph{Model-free reinforcement learning}
We learn the policy (actor) and action-value function (critic) using TD3 \citep{fujimotoAddressingFunctionApproximation2018}.
However, we follow \citet{yaratsMasteringVisualContinuous2021,zhaoSimplifiedTemporalConsistency2023}
and augment the loss with $n\text{-step}$ returns.
The only difference to TD3 is that instead of using the original observations $\vo_{t}$, we map them through the
online encoder $\vz_{t} = f(e_{{\theta}}(\vo_{t}))$ and learn the actor/critic in the quantized latent space $\vz_{t}$.
The critic is then updated by minimizing the following objective:
\begin{align} \label{eq:value-loss}
  \mathcal{L}_{q}(\psi; \tau) &= \E_{\tau \sim \mathcal{D}} \left[ \textstyle\sum_{k=1}^{2} (q_{\psi_{k}}(f(e_{{\theta}}(\vo_{t})), \va_{t}) - y)^{2}  \right], \quad  \forall k \in 1, 2 \\
  y &= \sum_{n=0}^{N-1} r_{t+n} + \gamma^{n} \min_{k \in \{1,2\}} q_{\bar{\psi}_{k}}(e_{{\theta}}(\vo_{t+n+1}), \va_{t+n+1}), \nonumber
\quad \text{with} \ \va_{t+n} = \pi_{\bar{\eta}}(\vz_{t+n}) + \epsilon_{t+n},
\end{align}
where we use policy smoothing by adding clipped Gaussian noise $\epsilon_{t+n} \sim \text{clip} \left(\mathcal{N} (0,\sigma^{2}), -c, c \right)$ to the
action $\va_{t+n} = \pi_{\bar{\eta}}(\vz_{t+n}) + \epsilon_{t+n}$.
Note that we use the online encoder to get the latent states in both the prediction and the target.
We then use the target action-value functions $\mathbf{q}_{\bar{\psi}}$ and the target policy $\pi_{\bar{\eta}}$ to
calculate the TD target.
Following TD3, we learn the actor's parameters by minimizing
\begin{align} \label{eq:policy-loss}
 \mathcal{L}_{\pi}(\eta ; \tau) = - \E_{\vo_{t} \sim \mathcal{D}} \bigg[ \min_{k\in\{1,2\}} q_{\psi_{k}}(\underbrace{f(e_{{\theta}}(\vo_{t}))}_{\vz_{t}}, \pi_{\eta}(f(e_{{\theta}}(\vo_{t})))) \bigg].
\end{align}
That is, we maximize the Q-value using the clipped double Q-learning trick to combat overestimation in Q-learning.
Note that we do not use the momentum encoder in the actor/critic objectives.
In our experiments, using the momentum encoder resulted in worse performance.

Whilst our method shares similarities with TCRL \citep{zhaoSimplifiedTemporalConsistency2023}, it is important to note that
our transition model does not predict the reward.
Instead, \our leverages quantization to help alleviate representation collapse, and, as a result,
learns a task-agnostic representation.

\section{Experiments}
\label{sec:experiments}
In this section, we evaluate \our in a variety of continuous control tasks from the DeepMind Control (DMC) Suite \cite{tassa2018deepmind}.
We aim to answer the following questions:
\begin{enumerate}
    \item How does \our compare to state-of-the-art model-free RL algorithms, especially in the hard DMC tasks?
    \item Does our FSQ-based quantization help combat representation and dimensional collapse?%
    \item Is learning a representation with only latent-state consistency really better than including reward predictions?
    \item What impact does reconstruction loss have on the performance of \our?
\end{enumerate}

\paragraph{\our is simple, fast, and performant}
We compare \our to the model-free baseline Twin Delayed DDPG (TD3, \cite{fujimotoAddressingFunctionApproximation2018}), and the representation learning-based RL methods Temporal Consistency Reinforcement Learning (TCRL, \cite{zhaoSimplifiedTemporalConsistency2023}), TD7 (\cite{fujimotoSALEStateActionRepresentation2023}), and Temporal Action-driven Contrastive Learning (TACO, \cite{zhengTextttTACOTemporal2023}).
In \cref{fig:normalization_improves_sample_efficiency}, we evaluate sample efficiency by plotting the average performance of the algorithms across 20 DeepMind Control Suite tasks as a function of environment steps.
We see that, on average, \our outperforms the baselines and shows significant advantages in many environments.
We outperform TCRL, which is the most similar baseline to our work.
Furthermore, TD3 is noncompetitive with \our, highlighting the importance of representation learning in state-based reinforcement learning. For complete results on all 20 tasks, see \cref{fig:dmc_grid} in Appendix~\ref{app:full_dmc}.
For more details of the tasks on which the algorithms were evaluated, see Appendix~\ref{app:tasks}.
For more details of the baselines used in our work and how we implemented them, see Appendix~\ref{app:baselines}.

\paragraph{High-dimensional control} Many tasks in DeepMind Control Suite are particularly high-dimensional. For instance, the observation space of the Dog tasks is $\mathcal{O} \in \R^{223}$ and the action space is $\mathcal{A} \in \R^{38}$, and for Humanoid, the observation space is $\mathcal{O} \in \R^{67}$ and the action space $\mathcal{A} \in \R^{24}$. \cref{fig:normalization_improves_sample_efficiency} and \cref{fig:dmc_grid} show that \our excels in the high dimensional Dog and Humanoid environments when compared to the baselines. We hypothesize that our discretized representations are particularly beneficial for simplifying learning the transition dynamics in high-dimensional spaces, making \our highly sample efficient in these tasks.

\paragraph{\our does not suffer from rank collapse}
We examine the behaviour of adding quantization to our MLP encoder during training.
Following \citet{niBridgingStateHistory2023}, we estimate the rank of the linear operator associated with
the MLP encoder by calculating the matrix rank\footnote{Rank of an $m\times n$ matrix $\mA$ is the dimension of the image of the mapping $g: \R^{n} \rightarrow \R^{m}$, with $g(\vx)=\mA \vx$} of the latent states for a batch of inputs.
We ensure full rank at the start of training by orthogonally initializing the MLP encoders.
\cref{fig:normalization-ablation} shows the orthogonality-preserving effect of our quantization scheme as
the matrix rank stays close to the maximum.
Without quantization, a dimensional collapse occurs, which can have significant harmful effects as the representational power of the latent state diminishes \cite{jingUnderstandingDimensionalCollapse2021}.
Correspondingly, in three of the four environments, removing the quantization has a deteriorating impact on the sample efficiency of \our, and in Dog Run, the algorithm completely fails to learn to solve the task without the quantization.

\begin{figure*}[t]
\vskip 0.2in
\begin{center}
\centerline{\includegraphics[width=\textwidth]{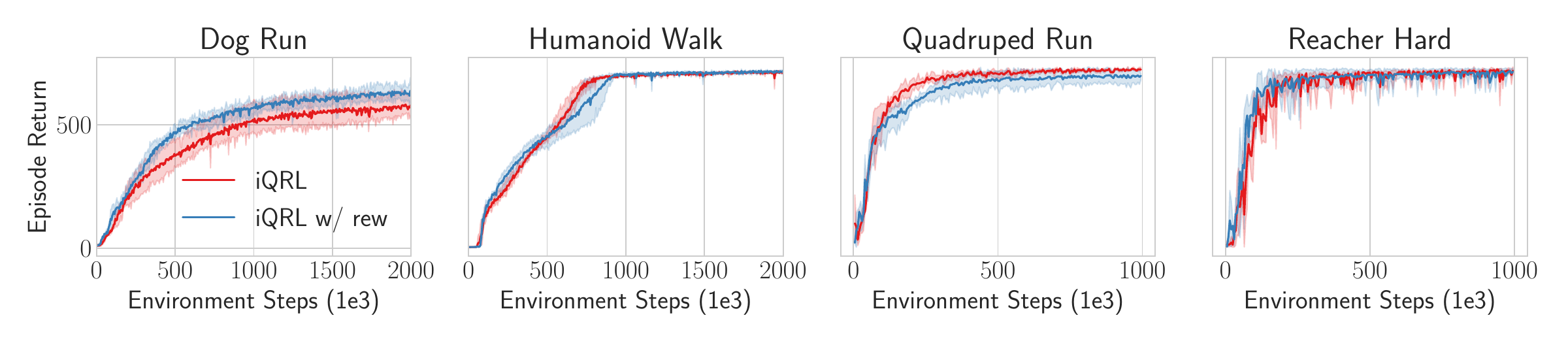}}
\caption{\textbf{Reward prediction is not necessary for representation learning.} We compare \our to a variant of our method with a reward prediction head trained to predict the reward from the current latent state. Adding a reward prediction head to \our leads into a slight increase in performance in Dog Run, but has a slightly harmful impact on sample efficiency in Humanoid Walk and Quadruped Run.  We plot the mean (solid line) and the $95\%$ confidence intervals (shaded) across 5 random seeds, where each seed averages over 10 evaluation episodes.}
\label{fig:reward-ablation}
\end{center}
\end{figure*}

\paragraph{Reward prediction is not necessary for representation learning}
Unlike prior methods such as TCRL \cite{zhaoSimplifiedTemporalConsistency2023}, TD-MPC \cite{hansenTemporalDifferenceLearning2022}, Dreamer-V2 \cite{hafnerMasteringAtariDiscrete2022} and TACO \cite{zheng2024texttt}, our representation learning loss (\cref{eq:rep-loss}) does not include a term for learning to predict the reward or the value of the latent state. Instead, we rely solely on the self-predictive temporal consistency loss. To analyze the impact of not including the reward prediction term, we compare our method to a variant of our method, where we have included a reward prediction head similar to that of \citet{zhaoSimplifiedTemporalConsistency2023}. Formally, we define a reward head as $\hat{r}_t = g_{\xi} (\vz_{t}, \va_{t})$ (see also \cref{eq:transition}), and include a reward prediction term (discounted MSE loss) in the representation loss:
	\begin{align*}
	\mathcal{L}_\text{rew} = \sum_{h=0}^{H-1} \gamma_{\text{rep}}^h \| \hat{r}_{t+h} - r_{t+h}\|_2^2,
\end{align*}
where $r_{t+h}$ is the ground-truth H-step reward and $\hat{r}_{t+h}$ is the predicted H-step reward.

The results for this ablation study are shown in \cref{fig:reward-ablation}. The plots show that our method, \our, without a reward prediction term in the loss, has equal or superior performance to the variant with a reward prediction term except in Dog Run. Our results imply that learning to predict the reward is not necessary for learning a suitable latent representation. The upside of not including the reward prediction head is that it makes the representation task-agnostic, which we believe to be important for downstream applications such as speeding up learning in an incremental multi-task setting in the same domain.\looseness-1

We also evaluated whether including a reward head alone without our FSQ-based normalization scheme is sufficient for preserving the rank of the latent representation and found that \our with a reward prediction head but without FSQ suffers from poor performance and dimensional collapse. Therefore, the reward prediction head is not a substitute for our quantization. For more details of this experiment, see \cref{app:further-ablations}.

\paragraph{Reconstruction loss has a detrimental impact}
Learning to minimize the observation reconstruction error has been widely applied in model-based RL \cite{sutton2018reinforcement, haRecurrentWorldModels2018, hafnerLearning2019}, and an observation decoder has been a component of many of the most successful RL algorithms to date \cite{hafner2023mastering}. However, recent work in representation learning for RL \citep{zhaoSimplifiedTemporalConsistency2023} and model-based RL \citep{hansenTemporalDifferenceLearning2022} has shown that incorporating a reconstruction term into the representation loss can hurt the performance, as learning to reconstruct the observations is inefficient due to the observations containing irrelevant details and visuals like shading that are uncontrollable by the agent and do not affect the tasks.

\begin{figure*}[t]
	\vskip 0.2in
	\begin{center}
		\centerline{\includegraphics[width=1.0\textwidth]{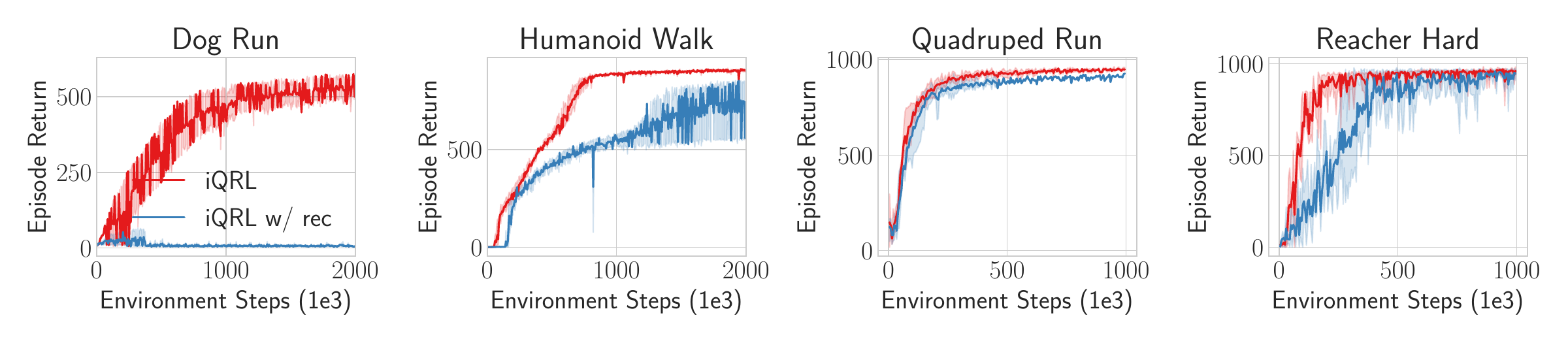}}
		\caption{\textbf{Reconstruction loss has detrimental impact.} Unlike many methods, such as SAC-AE \cite{yaratsImprovingSampleEfficiency2021}, \our neither has an observation decoder nor a reconstruction term in the loss function. We show that adding a reconstruction loss harms the performance of \our across a mixture of easy and hard evaluation environments. We plot the mean (solid line) and the $95\%$ confidence intervals (shaded) across 5 random seeds, where each seed averages over 10 evaluation episodes.}
		\label{fig:reconstruction-loss-ablation}
	\end{center}
\end{figure*}

To provide a thorough analysis of \our, we include results where we add a reconstruction term to our representation loss in \cref{eq:rep-loss}:
\begin{align}
	\mathcal{L}_{\vo} = \mathbb{E}_{\vo_t\sim\mathcal{D}} [\| \hat{\vo}_t - \vo_t \|_2^2], \quad
\hat{\vo}_t = h_\kappa(\vz_t),
\end{align}
where $h_\kappa$ is a learned observation decoder that takes the latent state as the input and outputs the reconstructed observation. The decoder $h_\kappa$ is a standard MLP. We perform reconstruction at each time step in the horizon.
The results in \cref{fig:reconstruction-loss-ablation} show that in no environments does reconstruction aid learning, and in some tasks, such as the difficult Dog Run and Humanoid Walk tasks, including the reconstruction term has a significant detrimental effect on the performance, and can even prevent learning completely.
Our results support the observations of \citet{zhaoSimplifiedTemporalConsistency2023} and \citet{hansenTemporalDifferenceLearning2022} about the lack of need for a reconstruction target in continuous control tasks.

\paragraph{Projection head}
\citet{wenMechanismPredictionHead2022} and \citet{schwarzerDataEfficientReinforcementLearning2020} investigated the role of a learnable projection head in non-contrastive self-supervised learning and found that it helps RL algorithms learn more diversified and therefore, superior representations. Whilst \our shares similarities with SPR \citep{schwarzerDataEfficientReinforcementLearning2020}, in particular, a temporal consistency loss using cosine similarity, it differs in that it does not use a learnable projection head and quantizes the
representation instead.
In \cref{fig:proj-ablation}, we show the impact of adding a projection head to \our.
It shows that the projection head decreases the sample efficiency of \our.
Whilst projection heads are effective for learning representations from images, our results suggest that they have
a significant negative impact on sample efficiency when learning representations of state-based observations, reaffirming that state-based RL has a different set of challenges to image-based RL and techniques designed to combat representation collapse are not always transferable between the settings.

\paragraph{Stop gradient}
\cite{niBridgingStateHistory2023} proved that using stop gradients should suffice for preventing representation collapse.
However, their experiments suggested that using an EMA encoder improves performance over simply using stop gradients.
In \cref{fig:stop-grad}, we show how replacing \our's EMA encoder with a stop gradient operation can have a
negative impact on performace.
For example, using stop gradient in the Acrobot Swingup task results in the agent struggling to solve the task.

\paragraph{Codebook size}
In \cref{app:codebook-ablation} we evaluate how the size of the codebook $|\mathcal{C}|$ influences the performance of the agent.
It shows that size of the codebook and the activeness is intuitive: the smaller the codebook the larger the active proportion.
The best codebook size varies between environments but the rank of the representation appears to be preserved for all codebook
sizes.

\paragraph{Latent dimension}
As a final experiment, we evaluate how the dimension of the latent space $d$ impacts
\our's performance. We find that \our is fairly robust to different latent dimensions.
We find that a latent dimension of $d=1024$ with FSQ levels $\mathcal{L}=[8,8]$, which corresponds to a codebook
size $|\mathcal{C}|=2^{6}$, performs best in the harder DMC tasks.
See \cref{app:latent-dim-ablation} for more details.

\section{Conclusion}
\label{conclusion}
We have presented \our, a technique for learning representations using only a self-supervised temporal consistency loss,
which demonstrates strong performance in continuous control tasks, including the complex DMC Humanoid and Dog tasks.
Our quantization of the latent space empirically preserves the representation's matrix rank,
indicating that it alleviates representation and dimensional collapse.
Our experiments further demonstrate that \our is extremely sample efficient whilst being fast to train,
which we believe is a strong selling point.
Importantly, our method is {\em (i)} straightforward, {\em (ii)} compatible with any model-free RL algorithm,
and {\em (iii)} learns a task-agnostic representation.

\paragraph{Limitations and future work}
Given that \our learns a task-agnostic representation, exploring its use for multi-task RL is an exciting direction for
future work.
Can \our learn a single representation which is shared across a wide variety of tasks?
In this paper, we have only evaluated \our in deterministic environments so extending \our to stochastic environments
is another important direction for future work.
\begin{ack}
AJS and KK were supported by the Research Council of Finland from the Flagship program: Finnish Center for Artificial Intelligence (FCAI).
YZ is funded by Research Council of Finland (grant id 345521) and MN is funded by Business Finland (BIOND4.0 - Data Driven Control for Bioprocesses).
AHS acknowledges funding from the Research Council of Finland (grant id 339730) and
JP acknowledges funding from Research Council of Finland (grant ids 345521 and 353198).
We acknowledge CSC -- IT Center for Science, Finland, for awarding this project access to the LUMI supercomputer, owned by the EuroHPC Joint Undertaking, hosted by CSC (Finland) and the LUMI consortium through CSC.
We acknowledge the computational resources provided by the Aalto Science-IT project.
\end{ack}

\phantomsection%
\addcontentsline{toc}{section}{References}
\begingroup
\small
\bibliographystyle{unsrtnat}

\endgroup

\clearpage
\appendix

\section*{Appendices}

\section{Implementation Details}

\paragraph{Architecture}
We implemented \our with PyTorch \citep{paszkePyTorchImperativeStyle2019} and used the AdamW
optimizer \citep{kingmaAdam2017} for training the models.
All components (encoder, dynamics, actor and critic) are implemented as MLPs.
Following \citet{hansenTDMPC2ScalableRobust2023} we let all intermediate layers be linear layers followed by
LayerNorm \cite{baLayerNormalization2016}.
Using LayerNorm is what led to our base TD3 implementation performing so well.
We use Mish activation functions throughout.
Below we summarize the \our architecture for our base model.

\begin{lstlisting}[basicstyle=\ttfamily\footnotesize]
iQRL(
  (fsq): FSQ(
    (project_in): Identity()
    (project_out): Identity()
  )
  (encoder): ModuleDict(
    (state): Sequential(
      (0): NormedLinear(in_features=O, out_features=256, act=Mish)
      (1): Linear(in_features=256, out_features=512)
    )
  )
  (encoder_tar): ModuleDict(
    (state): Sequential(
      (0): NormedLinear(in_features=O, out_features=256, act=Mish)
      (1): Linear(in_features=256, out_features=512)
    )
  )
  (dynamics): Sequential(
    (0): NormedLinear(in_features=512+A, out_features=512, act=Mish)
    (1): NormedLinear(in_features=512, out_features=512, act=Mish)
    (2): Linear(in_features=512, out_features=512)
  )
  (pi): Actor(
    (_pi): Sequential(
      (0): NormedLinear(in_features=512, out_features=512, act=Mish)
      (1): NormedLinear(in_features=512, out_features=512, act=Mish)
      (2): Linear(in_features=512, out_features=A)
    )
  )
  (pi_tar): Actor(
    (_pi): Sequential(
      (0): NormedLinear(in_features=512, out_features=512, act=Mish)
      (1): NormedLinear(in_features=512, out_features=512, act=Mish)
      (2): Linear(in_features=512, out_features=A)
    )
  )
  (critic): Critic(
    (_q1): Sequential(
      (0): NormedLinear(in_features=512+A, out_features=512, act=Mish)
      (1): NormedLinear(in_features=512, out_features=512, act=Mish)
      (2): Linear(in_features=512, out_features=1)
    )
    (_q2): Sequential(
      (0): NormedLinear(in_features=512+A, out_features=512, act=Mish)
      (1): NormedLinear(in_features=512, out_features=512, act=Mish)
      (2): Linear(in_features=512, out_features=1)
    )
  )
  (critic_tar): Critic(
    (_q1): Sequential(
      (0): NormedLinear(in_features=512+A, out_features=512, act=Mish)
      (1): NormedLinear(in_features=512, out_features=512, act=Mish)
      (2): Linear(in_features=512, out_features=1, bias=True)
    )
    (_q2): Sequential(
      (0): NormedLinear(in_features=512+A, out_features=512, act=Mish)
      (1): NormedLinear(in_features=512, out_features=512, act=Mish)
      (2): Linear(in_features=512, out_features=1)
    )
  )
)
\end{lstlisting}
where $O$ is the dimensionality of the observation space and $A$ is the dimensionality of the action spaces.

\paragraph{Hyperparameters}
\cref{tab:hyperparameters} lists all of the hyperparameters for training \our which were used for the main experiments
and the ablations.
\begin{table}[h]
\caption{\textbf{\our hyperparameters.} We kept most hyperparameters fixed across all tasks.}
\label{tab:hyperparameters}
\begin{center}
\resizebox{\textwidth}{!}{
\begin{footnotesize}
\begin{sc}
\begin{tabular}{lll}
\toprule
Hyperparameter & Value & Description \\
\midrule
\textbf{Training} & & \\
Action repeat & 2 & \\
Max episode length & 500 &  Action repeat makes this 1000 \\
Num. eval episodes & $10$ & \\
Random episodes & $10$ & Num. random episodes at start \\
\hline
\textbf{TD3} & & \\
Actor update freq. & $2$  & Update actor less than critic \\
Batch size & $256$ & \\
Buffer size & $10^{6}$ & \\
Discount factor $\gamma$ & $0.99$ & \\
Exploration noise & $\mathrm{Linear}(1.0,0.1,50)$ (easy) & \\
                  & $\mathrm{Linear}(1.0,0.1,150)$ (medium) & \\
                  & $\mathrm{Linear}(1.0,0.1,500)$ (hard) & \\
Learning rate & $3 \times 10^{-4}$ & \\
MLP dims & $[512, 512]$ & For actor/critic/dynamics \\
Momentum coef. ($\tau$) & $0.005$ & \\
Noise clip & $0.3$ & \\
N-step TD & $1$ or $3$ & \\
Policy noise & $0.2$ & \\
Update-to-data (UTD) ratio & $1$ & \\
\hline
\textbf{Encoder} &  & \\
Discount factor $\gamma_{\text{rep}}$ & $0.9$ & \\
Encoder learning rate & $10^{-4}$ & \\
Encoder MLP dims & $[256]$ & \\
Encoder momentum coef. ($\tau$) & 0.005 & \\
FSQ levels & $[8, 8]$ & \\
Horizon $(H)$ & $5$ & For representation learning \\
Latent dimension ($d$) & $512$  & \\
                       & $1024$ (Humanoid/Dog) & \\
 \bottomrule
\end{tabular}
\end{sc}
\end{footnotesize}
}
\end{center}
\vskip -0.1in
\end{table}

\paragraph{Statistical significance}
We used five seeds for the main figures, at least three seeds for all ablations, and plotted the 95 \% confidence intervals as the shaded area, which corresponds to approximately two standard errors of the mean.

\paragraph{Hardware}
We used Nvidia A100s and AMD Instinct MI250X GPUs to run our experiments. All our experiments have been run on a single GPU with a single-digit number of CPU workers.

\paragraph{Open-source code} For full details of the implementation, model architectures, and training, please check the code, which is available in the submitted supplementary material and will be made public upon acceptance to guarantee seamless reproducibility.

\vfill
\begin{center}
    --appendices continue on next page--
\end{center}

\clearpage
\section{Baselines}
\label{app:baselines}
In this section, we provide further details of the baselines we compare against.
In particular, we provide details of how we modified the original codebases and tuned the hyperparameters in an effort
to offer a fair comparison.

\begin{itemize}
\item \textbf{Temporal Consistency Reinforcement Learning (TCRL, \cite{zhaoSimplifiedTemporalConsistency2023})} is a reinforcement learning algorithm consisting of four components, an encoder and transition, policy and value functions, similarly to \our. TCRL uses a temporal consistency loss similar to model-based reinforcement learning to learn a representation used for model-free policy and value function training. The most crucial difference between TCRL and \our is that we replace the reward prediction head in the transition function with the FSQ-based normalization scheme. We used the official TCRL implementation on GitHub to run the TCRL experiments in our paper. For the DeepMind Control Suite (DMC) tasks, we used the tuned hyperparameters from the original paper. %
We used the official PyTorch implementation\footnote{\url{https://github.com/zhaoyi11/tcrl}}.
\item \textbf{Temporal Action-driven Contrastive Learning (TACO, \cite{zhengTextttTACOTemporal2023})} is a temporal contrastive learning framework that learns a latent representation of states and actions with a contrastive loss that optimizes the mutual information between the representations of current states and the following action sequences, and those of the corresponding future states. TACO was primarily designed for vision-based tasks, whereas our benchmarks are state-based. We adapted TACO to the state-based setting by increasing the learning rate and update-to-data ratios to match those of \our. We also replaced their CNN-based encoder with the MLP-based encoder of \our. Then, we performed a grid search over feature dimensions of 50 and 128, hidden dimensions of 256, 512, and 1024, and frame stacking and no frame stacking. We found the combination of a feature dimension of 50 and a hidden dimension of 1024 without frame stacking to perform the best.
\item \textbf{Twin Delayed DDPG (TD3, \cite{fujimotoAddressingFunctionApproximation2018})} is a model-free RL algorithm for continuous control, extending deep deterministic policy gradient (DDPG) to deal with value overestimation bias. Compared to DDPG, this algorithm uses two critics and takes the minimum over the two for training, adds clipped noise to the actions selected for bootstrapping (policy smoothing), and updates the actor less frequently compared to the critics. \our is based on TD3 and we simply replace the observations with their corresponding latent representation by mapping them through our encoder. This baseline uses our TD3 implementation which obtains very strong results. Comparing to this baseline allows us to investigate the impact of representation learning on sample-efficiency.

\item \textbf{TD7 \cite{fujimotoSALEStateActionRepresentation2023} } is a model-free reinforcement learning algorithm for continuous control that builds on TD3. TD7 builds on a representation learning method, state-action learned embeddings (SALE). The embeddings are learned using a temporal consistency term in the latent state. Other improvements that TD7 has over TD3 are prioritized experience replay and checkpointing. TD7 was initially evaluated on MuJoCo. To adapt it for DeepMind Control Suite, we added action repeats, essential for good performance on DMC. Then, we compared the original hyperparameters of TD7 to those of \our and found \our to perform the best, so we used those for the final evaluation. In particular, the exploration noise decay of \our was crucial for high performance in the DMC environments, and without it, TD7 struggled. Note that both TD7 and \our use TD3 as the underlying algorithm, allowing us to reliably compare the impact of SALE and our FSQ-based representations. We used the official PyTorch implementation of TD7\footnote{\url{https://github.com/sfujim/TD7}}.

\end{itemize}

\vfill
\begin{center}
    --appendices continue on next page--
\end{center}

\clearpage
\section{Tasks}
\label{app:tasks}
We evaluate our method in 20 tasks from the DeepMind Control suite \citep{tassa2018deepmind}.
\cref{tab:tasks} provides details of the environments we used, including the dimensionality of the observation
and action spaces.

\begin{table}[h]
\caption{\textbf{DMControl.} We consider a total of 20 continuous control tasks from the DeepMind Control suite.}
\label{tab:tasks}
\begin{center}
\begin{sc}
\begin{footnotesize}

\begin{tabular}{lccc}
\toprule
\textbf{Task} & \textbf{Observation dim} & \textbf{Action dim}  & \textbf{Sparse?} \\
\midrule
Acrobot Swingup & 6 & 1 & N \\
Cheetah Run & 17 & 6 & N \\
Cup Catch & 8 & 2 & Y \\
Dog Run & 223 & 38 & N \\
Dog Trot & 223 & 38 & N \\
Dog Stand & 223 & 38 & N \\
Dog Walk & 223 & 38 & N \\
Fish Swim & 24 & 5 & N \\
Hopper Hop & 15 & 4 & N \\
Hopper Stand & 15 & 4 & N \\
Humanoid Run & 67 & 24 & N \\
Humanoid Stand & 67 & 24 & N \\
Humanoid Walk & 67 & 24 & N \\
Quadruped Run & 78 & 12 & N \\
Quadruped Walk & 78 & 12 & N \\
Reacher Easy & 6 & 2 & Y \\
Reacher Hard & 6 & 2 & Y \\
Walker Run & 24 & 6 & N \\
Walker Stand  & 24 & 6 & N \\
Walker Walk  & 24 & 6 & N \\

 \bottomrule
\end{tabular}
\end{footnotesize}
\end{sc}
\end{center}
\vskip -0.1in
\end{table}

\vfill
\begin{center}
    --appendices continue on next page--
\end{center}

\clearpage
\section{Ablation of Codebook Size}
\label{app:codebook-ablation}

\begin{figure}[h]
	\begin{center}
			\centerline{\includegraphics[width=\textwidth]{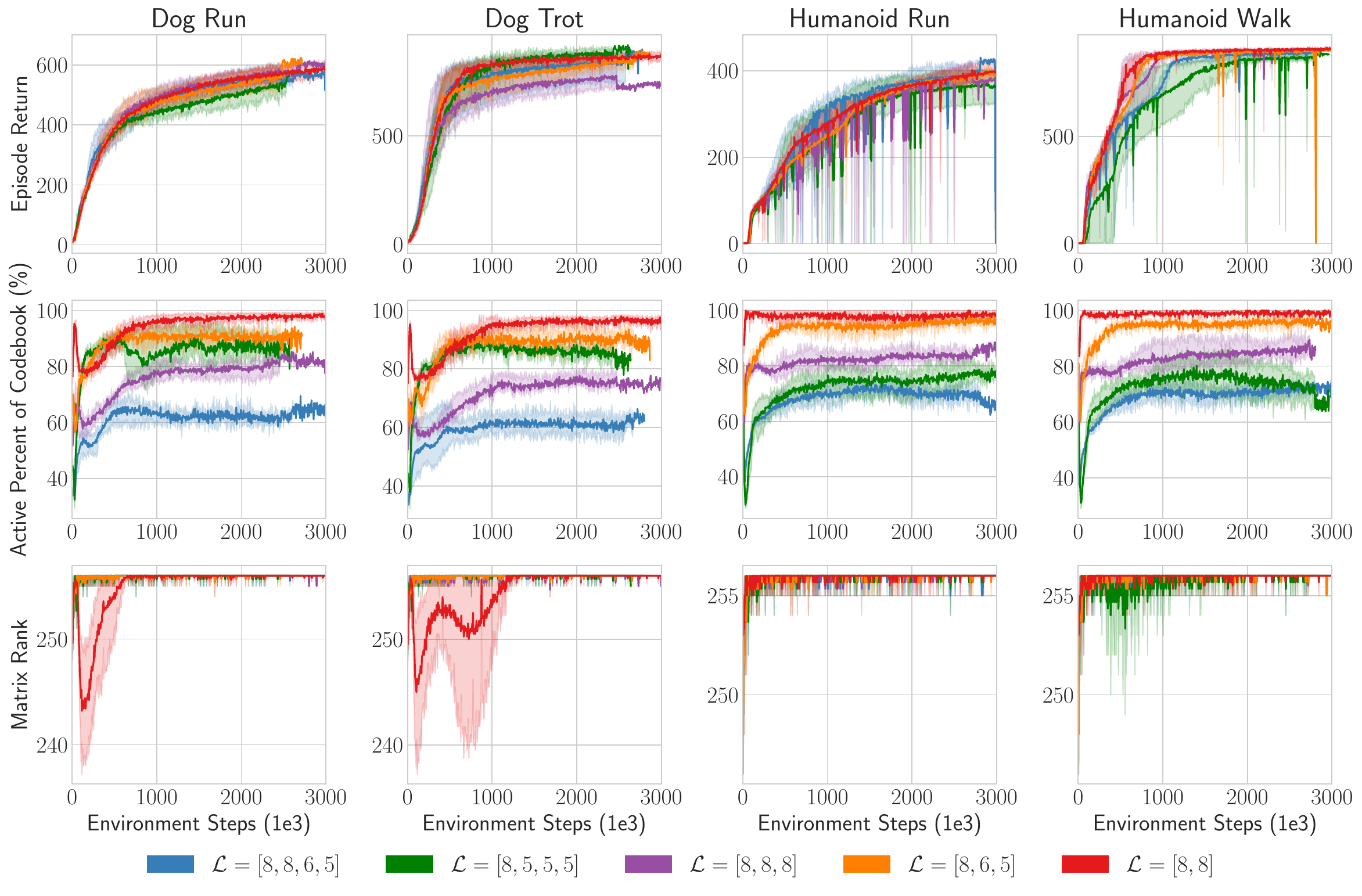}}
			\caption{\textbf{Codebook size ablation.} We compare how the codebook size affects the performance of \our (top), the percentage of the codebook that is active during training (middle), and how the different codebook sizes affect the encoder's ability to preserve the rank of the representation (bottom). In general, smaller codebooks become fully active faster than larger codebooks, and the rank of the representation is maintained for all codebook sizes. We plot the mean and the $95\%$ confidence intervals (shaded) across 3 random seeds for all environments.}
			\label{fig:codebook-size-ablation}
		\end{center}
\end{figure}

In this section, we evaluate how the size of the codebook $|\mathcal{C}|$ influences training.
We indirectly configure different codebook sizes via the FSQ levels $\mathcal{L} = \{L_{1},\ldots,L_{c} \}$ hyperparameter.
This is because the codebook size is given by $|\mathcal{C}| = \prod_{i=1}^{c} L_{i}$. The top row of \cref{fig:codebook-size-ablation} compares the training curves for different codebook sizes. The algorithm's performance is not particularly sensitive to the codebook size. A codebook that is too large can result in slower learning. The best codebook size varies between environments. The most difficult environment, Humanoid Run, benefits from the largest codebook.

Given that a codebook has a particular size, we can gain insights into how quickly \our's encoder starts to activate all of the codebook. The connection between the codebook size and the activeness of the codebook is intuitive: the middle row of \cref{fig:codebook-size-ablation} shows that the smaller the codebook, the larger the active proportion.

In the bottom row of \cref{fig:codebook-size-ablation}, we evaluate how different codebook sizes affect the encoder's ability to preserve the rank of the representation. We see that the rank of the representation is maintained no matter the codebook size.

\vfill
\begin{center}
    --appendices continue on next page--
\end{center}

\clearpage
\section{Ablation of Latent Dimension $d$}
\label{app:latent-dim-ablation}
\begin{figure}[h]
	\begin{center}
			\centerline{\includegraphics[width=\textwidth]{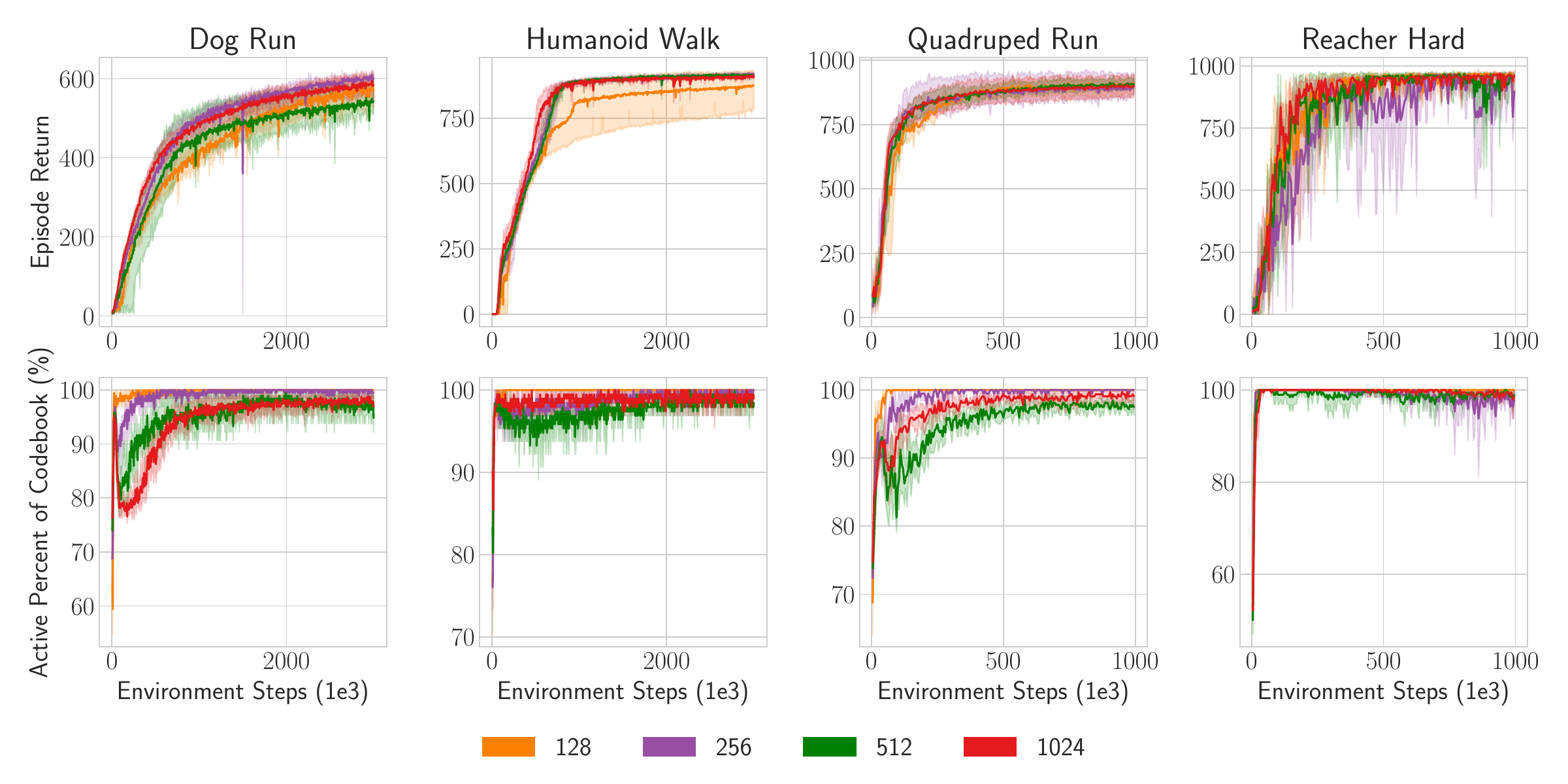}}
			\caption{\textbf{Latent dim $d$ ablation.} We compare how the latent dimension $d$ affects the performance of \our (top) and the percentage of the codebook that is active during training (bottom). In general, our algorithm is robust to the latent dimension of the representation, although in more difficult environments, such as Humanoid Walk, a $d$ too small can harm the agent's performance. We plot the mean and the $95\%$ confidence intervals (shaded) across 3 random seeds for all environments.}
			\label{fig:latent-dim-ablation}
		\end{center}
\end{figure}

This section investigates how the latent dimension $d$ affects the behavior and performance of \our in four different environments. The latent dimension $d$ corresponds to the dimension of the representation corresponding to each FSQ level before and after quantization is applied. In the top row of \cref{fig:latent-dim-ablation}, we see that the performance of our algorithm is robust to the latent dimension $d$, although a latent dimension too small can result in inferior performance, especially in the more difficult environments. The bottom row of \cref{fig:latent-dim-ablation} demonstrates that \our learns to use the complete codebook irrespective of the latent dimension. However, a larger $d$ can also correspond to the codebook becoming fully active slightly slower.

\vfill
\begin{center}
    --appendices continue on next page--
\end{center}

\newpage

\section{Further Ablations}
\label{app:further-ablations}

\begin{figure*}[h]
	\begin{center}
		\centerline{\includegraphics[width=\textwidth]{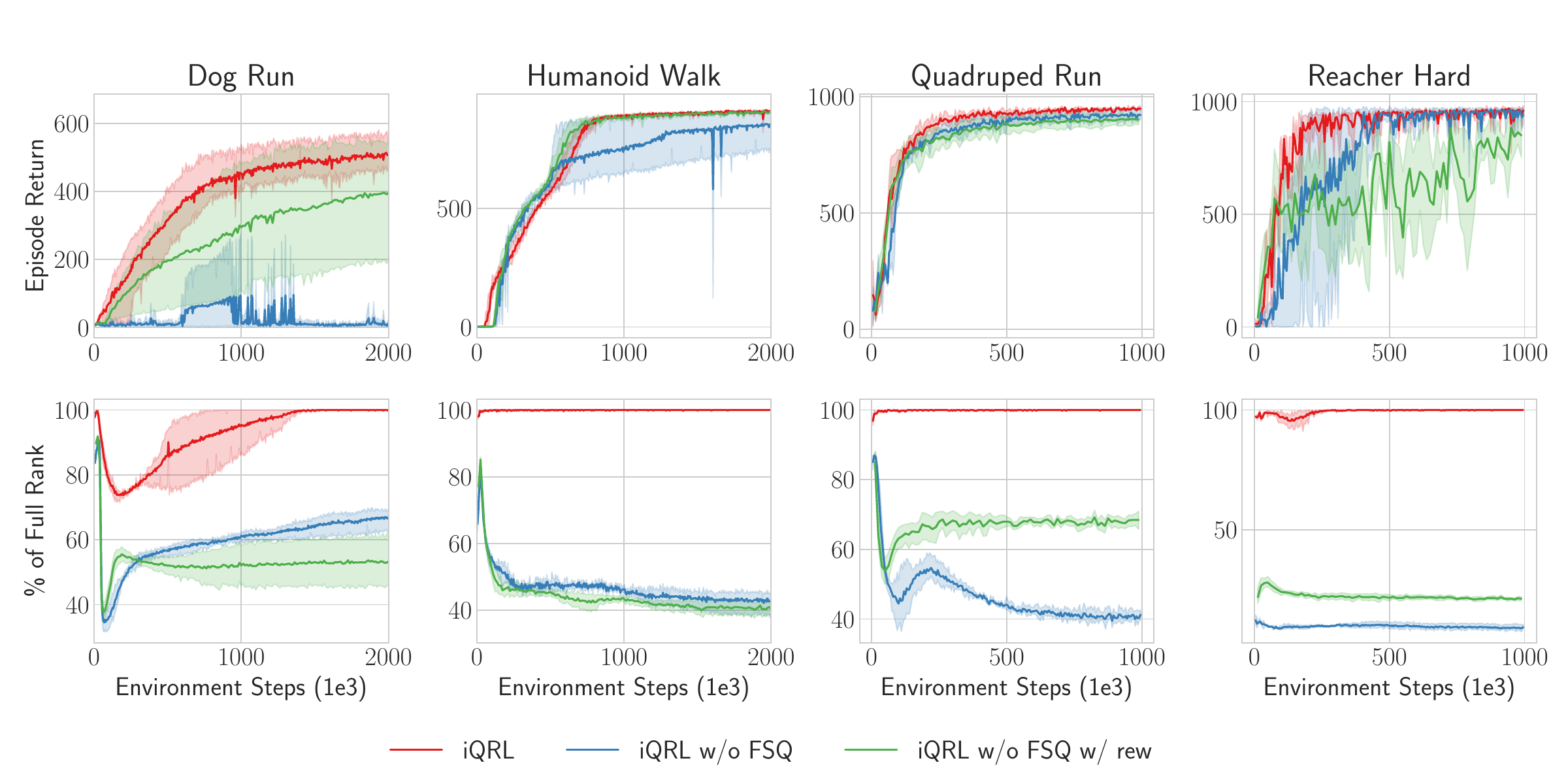}}
		\caption{\textbf{Adding a reward head is not enough to prevent loss of rank.} We show how removing our quantization scheme leads to dimensional collapse measured in terms of the rank of the representation, and in addition, how adding a reward prediction head to \our without quantization is insufficient to counteract this and maintain full rank. We plot the mean and the $95\%$ confidence intervals (shaded) across 3 random seeds for all environments.}
		\label{fig:rank-comparison-reward}
	\end{center}
	\vskip -0.2in
\end{figure*}

\begin{figure*}[h]
\begin{center}
\centerline{\includegraphics[width=\textwidth]{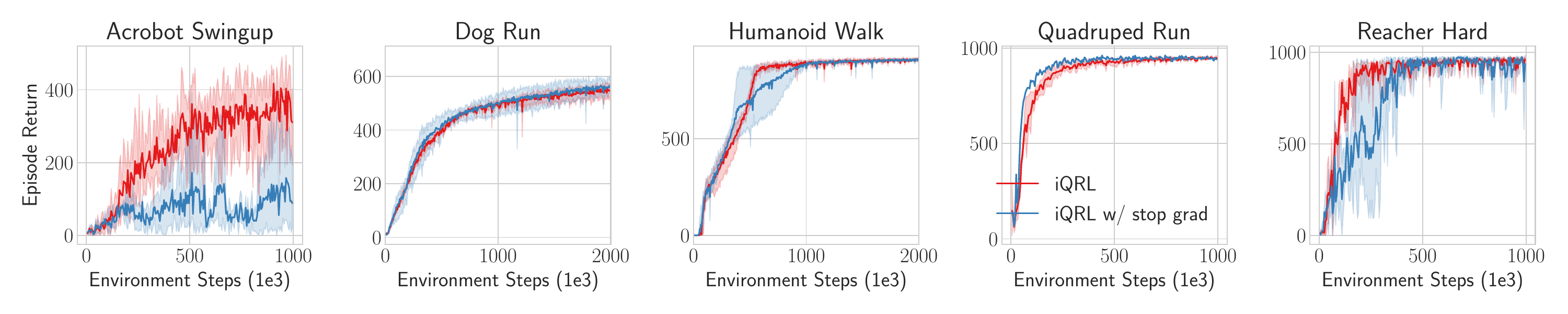}}
    \caption{\textbf{Replacing EMA encoder with stop gradient.} We show that removing \our's EMA encoder and replacing it with only stop gradient hurts performance in DMC tasks. This is particularly apparent in the Acrobot Swingup task. We plot the mean and the $95\%$ confidence intervals (shaded) across 3 random seeds for all environments.}
\label{fig:stop-grad}
\end{center}
\vskip -0.2in
\end{figure*}

\begin{figure*}[h]
\begin{center}
\centerline{\includegraphics[width=\textwidth]{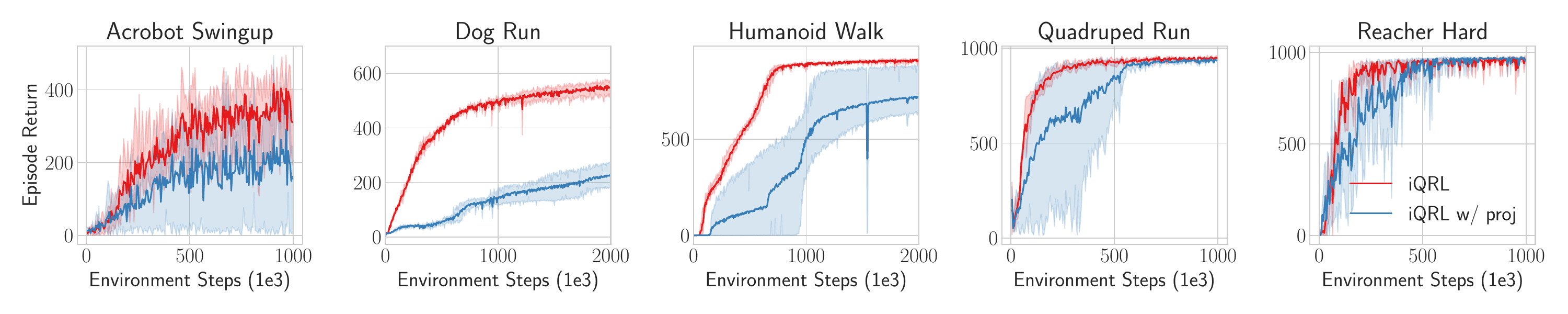}}
    \caption{\textbf{Adding a projection head decreases sample efficiency.} We show that adding a projection head to \our, similar to what is done in SPR \citep{schwarzerDataEfficientReinforcementLearning2020}, decreases \our's sample efficiency. We plot the mean and the $95\%$ confidence intervals (shaded) across 3 random seeds for all environments.}
\label{fig:proj-ablation}
\end{center}
\vskip -0.2in
\end{figure*}

\vfill
\begin{center}
    --appendices continue on next page--
\end{center}

\clearpage
\section{Further DMC Results}
\label{app:full_dmc}
\begin{figure*}[h]
\begin{center}
\centerline{\includegraphics[width=\textwidth]{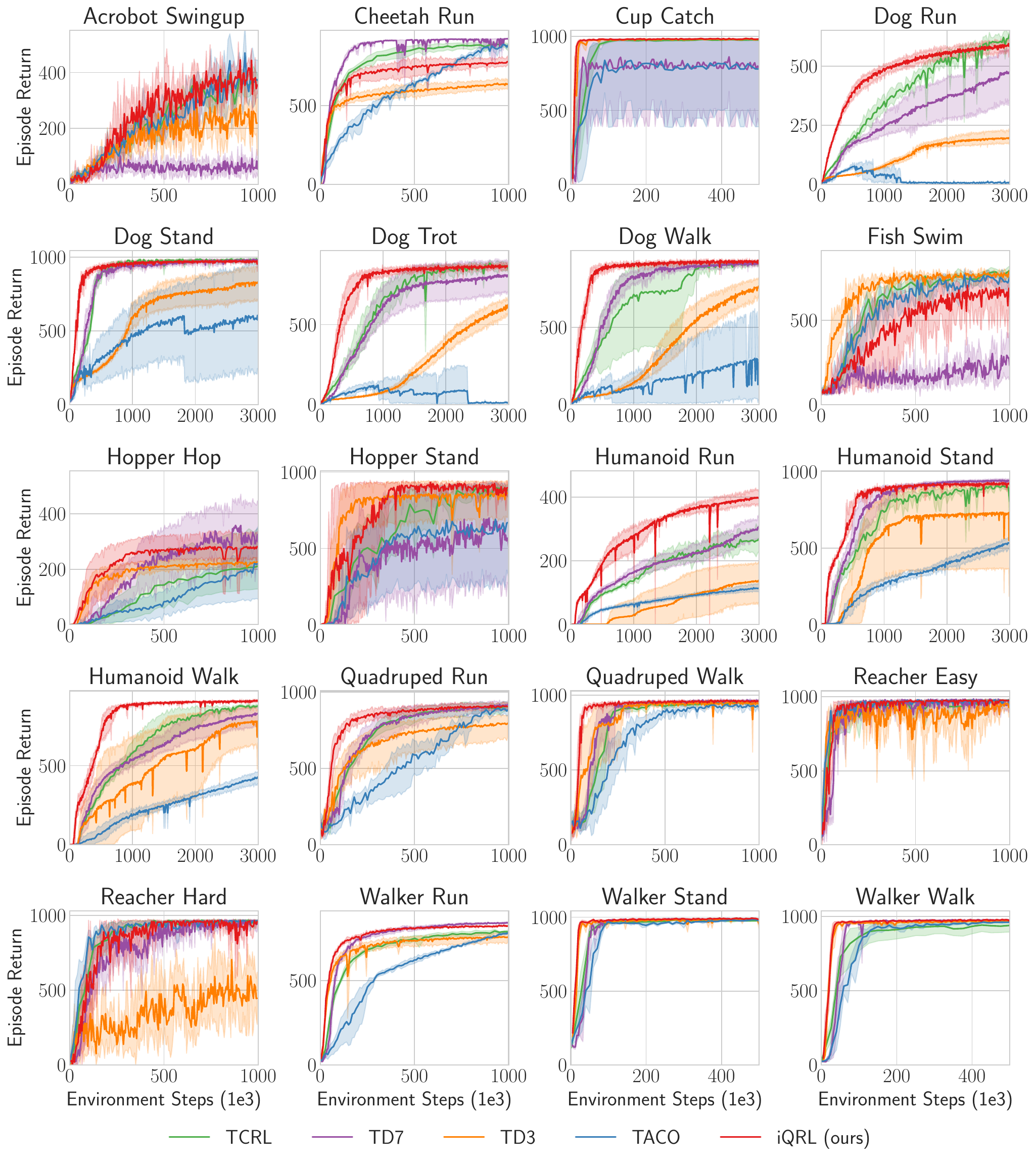}}
\caption{\textbf{DeepMind Control results.} \our performs well across a variety of DMC tasks. We plot the mean (solid line) and the $95\%$ confidence intervals (shaded) across 5 random seeds, where each seed averages over 10 evaluation episodes.}
\label{fig:dmc_grid}
\end{center}
\end{figure*}

\cref{fig:dmc_grid} compares \our to the baselines in the 20 DMC tasks. \our's representation
learning significantly improves sample efficiency when compared to TD3.
Note that \our uses the same TD3 implementation with the same hyperparameters, so the only difference is our
representation learning.
\our also outperforms TCRL in terms of sample efficiency, even without the reward prediction head.
Our experiments indicate that this improvement is due to our quantization and the inclusion of LayerNorm in our encoder.
We compare \our to TACO (which uses a contrastive loss) and observe that \our outperforms TACO in most environments.
TACO seems to particularly struggle in the Dog tasks.
Finally, \our outperforms TD7, a state-of-the-art representation learning method for state-based RL.

\end{document}